\documentclass[conference]{IEEEtran}
\IEEEoverridecommandlockouts
\usepackage{cite}
\usepackage{amsmath,amssymb,amsfonts}
\usepackage{algorithmic}
\usepackage{graphicx}
\usepackage{textcomp}
\usepackage{amssymb}
\usepackage{adjustbox}
\usepackage{multirow}
\usepackage{amsthm}
\newtheorem{theorem}{Theorem}
\usepackage{xcolor}
\usepackage{bm}
\usepackage{algorithm,algorithmic}
\usepackage{enumitem,hyperref}
\def\BibTeX{{\rm B\kern-.05em{\sc i\kern-.025em b}\kern-.08em
    T\kern-.1667em\lower.7ex\hbox{E}\kern-.125emX}}

\newtheorem{assumptionA}{Assumption}

\begin{document}

\title{Efficient Conformal Prediction for Regression Models Under Label Noise\\
{}
}

\author{\IEEEauthorblockN{Yahav Cohen}
\IEEEauthorblockA{\textit{Faculty of Engineering} \\
\textit{Bar-Ilan University}\\
Ramat Gan, Israel \\
coheny78@biu.ac.il}
\and
\IEEEauthorblockN{Jacob Goldberger}
\IEEEauthorblockA{\textit{Faculty of Engineering} \\
\textit{Bar-Ilan University}\\
Ramat Gan, Israel \\
jacob.goldberger@biu.ac.il}
\and
\IEEEauthorblockN{Tom Tirer}
\IEEEauthorblockA{\textit{Faculty of Engineering} \\
\textit{Bar-Ilan University}\\
Ramat Gan, Israel \\
tirer.tom@biu.ac.il}

}
\maketitle

\begin{abstract}
In high-stakes scenarios, such as medical imaging applications, it is critical to equip the predictions of a regression model with reliable confidence intervals. Recently, Conformal Prediction (CP) has emerged as a powerful statistical framework that, based on a labeled calibration set, generates intervals that include the true labels with a pre-specified probability.
In this paper, we address the problem of applying CP for regression models when the calibration set contains noisy labels.
We begin by establishing a mathematically grounded procedure for estimating the noise-free CP threshold. Then, we turn it into a practical algorithm that overcomes the challenges arising from the continuous nature of the regression problem. %
We evaluate the proposed method on two medical imaging regression datasets with Gaussian label noise. 
Our method %
significantly outperforms the existing alternative, achieving performance close to the clean-label setting.
\end{abstract}

\section{Introduction}
\label{sec:intro}

Regression deep neural networks (DNNs) play a pivotal role in modern machine learning, serving as the backbone of systems that predict continuous values from complex, high-dimensional data. They are widely applied in diverse domains: estimating patients' anatomical metrics from medical images, forecasting energy consumption and financial trends, guiding autonomous vehicles through trajectory prediction, and more.\cite{leibig2017leveraging,scher2018predicting,carvalho2015automated,avisdris2022}. 
For safety-critical applications, accurate prediction alone is not sufficient, and reporting the confidence of a prediction is essential for reliable and interpretable models. One widely adopted approach to conveying the uncertainty of a regression DNN on its predictions are confidence intervals, which should enclose the true value with a pre-defined probability. The size of these intervals is expected to be small and linked to the case’s complexity.

Recently, Conformal Prediction (CP) \cite{vovk1999machine,vovk2005conformal,angelopoulos2023conformal} has emerged as a powerful general statistical framework for this purpose.
Based on a labeled calibration set, CP returns a confidence interval (or ``prediction set'' in the case of classification) with a coverage guarantee, under very mild assumptions on the data distribution (i.i.d.~or, more generally, exchangeability of the calibration and test samples). 
Specifically, it is guaranteed that the confidence interval includes the true label with a pre-specified probability.
The goal is to return the smallest interval while maintaining the coverage level, and different CP methods are judged according to the average length of their confidence interval or ``prediction set'' \cite{angelopoulos2023conformal,angelopoulos2020uncertainty,dabah2024temperature,fargion2025enhancing}, referred to in the literature also as \textit{efficiency}.
CP has become an important calibration tool in safety-critical applications such as medical imaging \cite{lu2022improving,lu2022fair,olsson2022estimating,cosarinsky2026confic}. 
Up-to-date reviews of CP applied to regression are \cite{kato2023review,zhou2025conformal}. 

A critical challenge for CP in applications such as medical imaging arises from label noise. In these domains, datasets frequently contain noisy labels stemming from ambiguous data that can confuse even clinical experts. Furthermore, physicians may disagree on the diagnosis for the same medical image, leading to inconsistencies in the ground truth labeling. %
The challenge of calibrating the models based on noisy labels has only recently begun to receive attention, mostly for the classification setup. Einbinder et al.~\cite{einbinder2024label} suggested ignoring label noise and simply applying the standard CP algorithm on the noisy labeled calibration set. For regression, this strategy results in larger confidence intervals.  
Other related works
\cite{sesia2023adaptive,Penso_2024,penso2025conformal,Clarkson2024} offer adaptations of CP to noisy labels but focus on classification.
Some of them present coverage guarantee bounds. Yet, these bounds are too conservative in many cases (leading to very large prediction sets).

In this study, we are the first to tackle the challenge of applying CP to regression DNNs using calibration datasets with noisy labels.
We present a novel algorithm for estimating the noise-free CP threshold, developed by establishing a mathematically grounded procedure and then turning it into a practical method that overcomes the challenges arising from the continuous nature of the regression problem. 
We also discuss the case where no clean labels are available during model training.
We evaluate the proposed method on two medical imaging regression datasets with Gaussian label noise. 
Our method demonstrates robustness to the noise level, effective coverage, and yields average interval length (efficiency) close to the clean-label setting, significantly shorter than that obtained by the approach of \cite{einbinder2024label}, which motivates ignoring the label noise.

\section{Background}
\label{sec:background}

Let $(X, Y)$ denote a sample and its label, distributed over $\mathcal{X} \times \mathcal{Y}$. 
For a regression task with scalar label, we have $\mathcal{Y} = \mathbb{R}$.
Consider a DNN that for an input sample $x \in \mathcal{X}$ outputs a prediction $\hat{y}(x) \in \mathcal{Y}$, a calibration set of labeled samples $\{x_i,y_i\}_{i=1}^n$, and a predefined $\alpha \in (0,1)$.

Conformal Prediction (CP) establishes a decision rule for generating confidence intervals $x \mapsto \mathcal{C}(x)$, such that $Y \in \mathcal{C}(X)$ with probability $1-\alpha$, where $Y$ is the true class associated with $X$ \cite{vovk1999machine, vovk2005conformal}.
The only assumption in CP is that the random variables associated with the calibration set and the test samples are exchangeable (e.g., the samples are i.i.d.).
Let us state the general CP framework  \cite{angelopoulos2023conformal}:
\begin{enumerate}[leftmargin=*]
    \item Define a heuristic score function $s(x,y) \in \mathbb{R}$ based on some output of the model. A higher score should encode a lower level of agreement between $x$ and $y$.
    \item Calibration phase: Compute $\hat{q}$ as the $\lceil (n+1)(1-\alpha) \rceil/{n}$ quantile of the scores $\{s(x_1,y_1),\dots , s(x_n,y_n)\}$.
    \item Deployment phase: Use $\hat{q}$ to create the prediction set for a new sample $x_{n+1}$:
    $\mathcal{C}(x_{n+1}) = \{y:\,s(x_{n+1},y) \leq \hat{q} \}$.
\end{enumerate}

In the regression case, where $\mathcal{Y} = \mathbb{R}$, essentially we get the confidence interval $C(x_{n+1}) \subset \mathbb{R}$.
CP methods possess the following coverage guarantee. 

\begin{theorem}[Theorem 1 in \cite{angelopoulos2023conformal}]
\label{thm:cp_guarantees}
Suppose that $\{\left(X_i, Y_i\right)\}_{i=1}^n$ and $(X_{n+1},Y_{n+1})$ %
are i.i.d. 
Define $\hat{q}$ as in step 2 above and $\mathcal{C}(X_{n+1})$ as in step 3 above. Then the following holds: 
    $\mathbb{P}\left(Y_{n+1} \in \mathcal{C}(X_{n+1})\right) \geq 1 - \alpha$.
\end{theorem}


The proof of this result is based on \cite{vovk1999machine}.
A proof of an upper bound of $1-\alpha+1/(n+1)$ also exists.

Different CP methods typically differ by their choice of score function $s(x,y)$, and a key property that they are judged according to is the average $|\mathcal{C}(x)|$ (length or cardinality of $\mathcal{C}(x)$), often termed \textit{efficiency}.
Focusing on regression, the simplest choice of $s(x,y)$ is:
$
s(x,y) = |y-\hat{y}(x)|.
$
However, this yields an interval of fixed size for any sample since: $\mathcal{C}(x) = \{y:\, |y-\hat{y}(x)|\leq \hat{q} \} \implies |\mathcal{C}(x)|=2\hat{q}$, thereby ignoring whether $x$ is an easy or hard sample.
Note that the output $\hat{y}(\cdot)$ of the regression DNN can be understood as an estimate of the posterior's mean.
It is common to train the DNN to output also $\hat{u}(\cdot)$, an estimate of the posterior's standard deviation, e.g., using the Gaussian negative-log-likelihood (NLL) loss (in practice, outputting $\log(\hat{u}(\cdot))$ facilitates optimization).
In this case, a score function that yields better efficiency is given by
$
s(x,y) = |y-\hat{y}(x)|/\hat{u}(x),
$
and its confidence interval is $C(x) = \big[\hat{y}(x) - \hat{q}\hat{u}(x), \hat{y}(x) + \hat{q}\hat{u}(x)\big]$ \cite{nizhar2025clinical}.
For conciseness, this score and resulting $C(x)$ will also be used in this paper.

In this paper, we consider the problem of having a calibration set $\{x_i,\tilde{y}_i\}_{i=1}^n$, where $\tilde{y}_i$ is a noisy version of $y_i$.
This case has been studied in \cite{einbinder2024label}.
The authors have shown that for dispersive noise, where the distribution of $\tilde{Y}|X$ is more spread than the distribution of $Y|X$, performing the regular calibration phase of CP on $\{x_i,\tilde{y}_i\}_{i=1}^n$ will lead to threshold $\hat{q}$ that is larger than the one obtained from noise-free calibration, and hence coverage is maintained: $\mathbb{P}\left(Y_{n+1} \in \mathcal{C}(X_{n+1})\right) \geq 1 - \alpha$, where $Y_{n+1}$ is the clean label of the test sample $X_{n+1}$.

Unfortunately, as was shown in previous works for classification  \cite{sesia2023adaptive,Penso_2024,penso2025conformal,Clarkson2024} and will be shown here for regression, this simple approach, dubbed ``noisy CP'', leads to significant over-coverage (more than the desired $1-\alpha$) and consequently also to very large confidence intervals (poor efficiency).

\section{Conformal Prediction for Regression with Label Noise}
\label{sec:method}

The goal of this paper is to design a CP method that satisfies the pre-defined coverage requirement while providing confidence intervals that are much shorter than those of the ``noisy CP''.
To this end, we will base our method on estimating the CP threshold of the noise-free case.

\subsection{Estimating the noise-free CP threshold}

Let $X$, $Y$ and $\tilde{Y}$ be continuous random variables denoting the sample, its clean label and its noisy label, respectively.
Let $C_q(x)$ be a confidence interval depending on a threshold $q$, e.g., $C_q(x) = [ \hat{y}(x)-q\hat{u}(x) , \hat{y}(x)+q\hat{u}(x) ]$, associated with the score $s(x,y) = |y-\hat{y}(x)|/\hat{u}(x)$, for a model that outputs the prediction $\hat{y}(\cdot)$ and heuristic uncertainty $\hat{u}(\cdot)$.
We make the following assumptions.

\begin{assumptionA}
    \label{assump:independence}
    $\tilde{Y}$ and $X$ are conditionally independent given $Y$.
\end{assumptionA}

\begin{assumptionA}
    \label{assump:kernel}
    $p_{\tilde{Y}|Y}(\tilde{y}|y)=k(\tilde{y}-y;\sigma^2)$, i.e., the noise can be expressed as a kernel $k$ with  bandwidth parameter $\sigma^2$.
\end{assumptionA}

Assumption \ref{assump:independence} is common in the literature on label noise \cite{sesia2023adaptive,Penso_2024,penso2025conformal,Clarkson2024}.
Assumption \ref{assump:kernel} includes, for example, the case of additive Gaussian noise: $k(\cdot;\sigma^2) = \mathcal{N}(\cdot;0,\sigma^2)$.

Define
\begin{align*}
    M_q^c(\ell,y) &:= \mathbb{P}(\ell \in C_q(X)|Y=y) p_Y(y), \\
    M_q^n(\ell,\tilde{y}) &:= \mathbb{P}(\ell \in C_q(X)|\tilde{Y}=\tilde{y}) p_{\tilde{Y}}(\tilde{y}).
\end{align*}
Observe that
\begin{align}
\label{eq:Pcov_via_Mcq}
    \mathbb{P}(Y \in C_q(X)) &= \int \mathbb{P}(\ell \in C_q(X)|Y=\ell) p_Y(\ell) \mathrm{d}\ell \nonumber \\ 
    &= \int M_q^c(\ell,\ell) \mathrm{d}\ell.
\end{align}
That is, the probability of coverage of the true label depends on the ``trace'' of $M_q^c$. However, from a noisy labeled calibration set $\{x_i,\tilde{y}_i\}_{i=1}^n$ we may only approximate $M_q^n$.
This motivates us to establish a relation between $M_q^n$ and $M_q^c$. Indeed, we have the following:
\begin{align}
\label{eq:Mcq_via_deconv}
    &M_q^n(\ell,\tilde{y}) = \mathbb{P}(\ell \in C_q(X)|\tilde{Y}=\tilde{y}) p_{\tilde{Y}}(\tilde{y}) \nonumber \\
    &= \int \mathbb{P}(\ell \in C_q(X)|\tilde{Y}=\tilde{y},Y=y) p_{Y|\tilde{Y}}(y|\tilde{y}) p_{\tilde{Y}}(\tilde{y}) \mathrm{d}y \nonumber \\
    &= \int \mathbb{P}(\ell \in C_q(X)|\tilde{Y}=\tilde{y},Y=y) p_{Y}(y) p_{\tilde{Y}|Y}(\tilde{y}|y) \mathrm{d}y \nonumber \\
    &= \int \mathbb{P}(\ell \in C_q(X)|Y=y) p_{Y}(y) p_{\tilde{Y}|Y}(\tilde{y}|y) \mathrm{d}y \nonumber \\  
    &= \int M_q^c(\ell,y) p_{\tilde{Y}|Y}(\tilde{y}|y) \mathrm{d}y  \nonumber \\
    &=\left [ M_q^c(\ell,\cdot) * k(\cdot;\sigma^2) \right ] (\tilde{y})
\end{align}
where the third equality follows from Bayes rule, the fourth equality uses %
\ref{assump:independence},
and the last equality uses \ref{assump:kernel} with `$*$' denoting the convolution operation.

Using the above results, we can design a procedure for estimating the CP threshold $\hat{q}$ of the noise-free case. Specifically, given a value of $q$, we can compute (approximate) $M_q^n(\ell,y)$ for $\ell,\tilde{y} \in \mathcal{Y}$ using the noisy calibration data.
Then, \eqref{eq:Mcq_via_deconv} shows that to obtain (approximate) $M_q^c(\ell,\cdot)$ per $\ell \in \mathcal{Y}$, we need to solve a 1D deconvolution problem with $M_q^n(\ell,\cdot)$ and the noise kernel $k$.
Next, using the relation in \eqref{eq:Pcov_via_Mcq}, we can approximate the coverage probability $\mathbb{P}(Y \in C_q(X))$.
If it is larger (resp.~smaller) than $1-\alpha$, we need to slightly decrease (resp.~increase) $q$ and repeat the procedure.
This conceptual strategy is presented in Algorithm~\ref{alg:high_level_alg}, where we utilize the fact that our noise model is dispersive, and thus initializing $q$ using the ``noisy CP'' approach will lead to a large value that we can gradually reduce until reaching a stopping criterion $\mathbb{P}(Y \in C_q(X)) \gtrapprox 1-\alpha$.

\begin{algorithm}[t]
   \caption{Conceptual high-level procedure for computing $\hat{q}$ (see Section~\ref{sec:practical} for practical implementation)}
 \begin{algorithmic}
   \STATE {\bfseries Input:} $\alpha$, small $\delta_q$ (default: 0.05), $\mathcal{C}_q(\cdot)$ prediction interval rule depending on threshold $q$, calibration set with noisy labels $\{x_i,\tilde{y}_i\}_{i=1}^n$, noise kernel $k(\cdot;\sigma^2)$. 
   \STATE \textbf{Initialize}: $q \gets $ plain CP calibration using $\{x_i,\tilde{y}_i\}_{i=1}^n$
   \REPEAT
   \STATE $q \gets q - \delta_q$ 
   \STATE \textcolor{blue}{\# Step 1: Compute $M_q^n(\ell,\tilde{y})$ using the noisy labels}
    \STATE For $\ell,\tilde{y} \in \mathcal{Y}$: $M_q^n(\ell,\tilde{y}) \gets \mathbb{P}(\ell \in C_q(X)|\tilde{Y}=\tilde{y}) p_{\tilde{Y}}(\tilde{y})$
   \STATE \textcolor{blue}{\# Step 2: Compute $M_q^c(\ell,\tilde{y})$ using deconvolution}
    \STATE For $\ell \in \mathcal{Y}$: $M_q^c(\ell,\cdot) \gets$ deconv. $M_q^n(\ell,\cdot)$ and $k(\cdot;\sigma)$    
   \STATE \textcolor{blue}{\# Step 3: Compute coverage probability}
    \STATE $\mathbb{P}(Y \in C_q(X)) \gets \int M_q^c(\ell,\ell) \mathrm{d}\ell$
   \UNTIL{$\mathbb{P}(Y \in C_q(X)) < 1-\alpha$}
    \STATE \textbf{return} $\hat{q} \gets q+\delta_q$ 
 \end{algorithmic}
 \label{alg:high_level_alg}

\end{algorithm}

\subsection{Practical implementation}
\label{sec:practical}

Let us explain how we implement in practice each step in Algorithm~\ref{alg:high_level_alg}.
Clearly, the main challenge is that the probability and distribution that appear in the definition of $M_q^n$ are unknown and we can only approximate them from finite samples $\{x_i,\tilde{y}_i\}_{i=1}^n$. 
This will necessitate using discretizations.
Yet, despite our approximations, in Section~\ref{sec:exp} we will show the strong empirical performance of our method.

\noindent
\textbf{Discretization.}
The domain $\mathcal{Y}$ is discretized by partitioning a range slightly larger than $[\min_i \tilde{y}_i,  \max_i \tilde{y}_i]$ into bins $\{B_{\ell}\}$ of width $\delta_y=0.01$. Denote by $L$ the number of bins. From now we use $\ell,\tilde{y}$ to denote indices.
Each noisy label $\tilde{y}_i$ is assigned to a bin $B_{\ell}$, together with the corresponding input $x_i$.

\noindent
\textbf{Step 1.}
For a given $q>0$, 
we approximate $M_q^n$ by an $L\times L$ matrix $\hat{M}_q^n$ computed as follows:
$\hat{M}_q^n[\ell, \tilde{y}] = \frac{\sum_i \mathbb{I}\{B_{\ell} \subseteq C_q(x_i) \cap \tilde{y}_i \in B_{\tilde{y}} \}}{|\tilde{y}_i: \tilde{y}_i \in B_{\tilde{y}} |} 
 \cdot \frac{1}{\delta_y} \frac{\sum_i \mathbb{I}\{ \tilde{y}_i \in B_{\tilde{y}}\}}{n}$.
That is, we approximate the probabilities by their empirical versions.

\noindent
\textbf{Step 2.}
Given a discretization of the kernel, $\hat{k}$, we approximate $M_q^c$ by solving the convex optimization problem
\begin{align*}
\min_{\substack{\hat{M}_q^c \in \mathbb{R}^{L \times L} \\ 0 \le \hat{M}_q^c[\ell,\tilde{y}] \le 1/\delta_y}} \; 
& \sum_{\ell} \Big\| (\hat{M}_q^c[\ell,:] \ast \hat{k} - \hat{M}_q^n[\ell,:] ) 
\odot \mathbf{1}_{\hat{M}_q^n[\ell,:] > \epsilon} \Big\|_2^2 \nonumber \\
& + \lambda \, \| \hat{M}_q^c \|_F^2
\end{align*}
where $\odot$ denotes element-wise multiplication that we use to mask out bins that are empty in $\hat{M}_q^n[\ell,:]$. We found that this masking is necessary for handling the fact that the calibration set is finite. 
We set the regularization parameter to $\lambda = 0.01$. 

\noindent
\textbf{Step 3.}
We use $\sum_{\ell} \hat{M}_q^c[\ell,\ell]\delta_y$ as an approximation of the coverage probability.

\subsection{Estimating the level of label noise}
\label{seq:noise_est}

Algorithm~\ref{alg:high_level_alg} requires knowledge of the noise kernel $k(\cdot;\sigma)$.
In particular, under a Gaussian noise assumption $\tilde{Y}|Y \sim \mathcal{N}(0,\sigma^2)$, the variance $\sigma^2$ needs to be known or estimated.
Let us present a simple estimator for $\sigma^2$, given a model that has been trained with noisy labels. %
Future work may further investigate the problem of estimating the label noise model.

We begin with the relation
\begin{align*}
p_{\tilde{Y}|X}(\tilde{y}|x) &= \int p_{\tilde{Y}|Y,X}(\tilde{y}|y,x) p_{Y|X}(y|x)  dy \\
&=\int p_{\tilde{Y}|Y}(\tilde{y}|y) p_{Y|X}(y|x)  dy,
\end{align*}
where the second equality uses \ref{assump:independence}.
Assuming that $Y|X \sim \mathcal{N}(\mu(X),{u}^2(X))$, 
together with $\tilde{Y}|Y \sim \mathcal{N}(0,\sigma^2)$, the integral is essentially convolution of Gaussians that gives $\tilde{Y}|X \sim \mathcal{N}(\mu(X),{u}^2(X) + \sigma^2)$. 

Since the model $(\hat{y}(\cdot),\hat{u}^2(\cdot))$ is trained with Gaussian NLL, for a sample $x$, we estimate ${u}^2(x) + \sigma^2$ simply by $\hat{u}^2(x)$.
Therefore, under the mild assumption of having ``easy'' samples $\{x_i\}$ for which ${u}^2(x_i) \approx 0$,
$\sigma^2$ can be estimated from the smallest values of $\hat{u}^2(\cdot)$ across the training set.
In our experiments, we observe that the average of the lowest $1\%$ values of $\hat{u}^2(\cdot)$ provides a satisfactory prediction. %

\begin{table*}[t]
    \caption{Coverage and interval length for $\alpha$=0.1 and true $\sigma_{\text{true}}$=0.2. Our method is applied also with ``wrong'' $\sigma$ values.}
    \vspace{1mm}
    \centering
     \scalebox{0.85}{ 
    \begin{tabular}{|c|l|c|c|c|}
        \hline
        \textbf{Dataset} & \textbf{Method} & \textbf{$\hat{q}$ value} & \textbf{Avg.~length $\downarrow$}  & \textbf{Coverage(\%)}\\
        \hline
        \multirow{5}{*}{Chest X-Ray} 
            & Oracle CP   & $2.48 \pm 0.01 $  & $0.58 \pm 0.00 $ & $90.02 \pm 0.00 $  \\
            & Noisy CP    & $3.73 \pm 0.01 $ & $0.95 \pm 0.01 $ & $96.10 \pm 0.30 $  \\
            & Ours w/ $\sigma$=0.15  & $2.71 \pm 0.02 $  & $0.71 \pm 0.02 $ & $91.45 \pm 0.15 $ \\
            & Ours w/ $\sigma$=0.2 & $2.53 \pm 0.02 $ & $0.67 \pm 0.01 $ & $90.57 \pm 0.22 $  \\
            & Ours w/ $\sigma$=0.25 & $2.36 \pm 0.02 $  & $0.60\pm 0.01 $  & $88.52 \pm 0.16 $ \\
        \hline
        \multirow{5}{*}{BoneAge} 
            & Oracle CP   & $2.40 \pm 0.02 $  & $0.66 \pm 0.01 $ & $90.10 \pm 0.00 $  \\
            & Noisy CP    & $3.31 \pm 0.03 $ & $0.94 \pm 0.01 $ & $94.14 \pm 0.30 $  \\
            & Ours w/ $\sigma$=0.15  & $2.44 \pm 0.02 $  & $0.79 \pm 0.01 $ & $90.84 \pm 0.15 $ \\
            & Ours w/ $\sigma$=0.2 & $2.42 \pm 0.02 $ & $0.74 \pm 0.02 $ & $90.11 \pm 0.14 $  \\
            & Ours w/ $\sigma$=0.25 & $2.37 \pm 0.02 $  & $0.60\pm 0.01 $  & $89.72 \pm 0.17 $ \\
        \hline
    \end{tabular}
    }
    \label{tab:coverage_ci}
\end{table*}

\begin{table*}[t]
    \caption{Coverage and interval length for the Chest X-Ray dataset, $\alpha$=0.1 and $\sigma_{\text{true}}\in \{0.2,0.3 \}$. Our method is applied with an estimate $\hat{\sigma}$.}
    \vspace{1mm}
    \centering
    \scalebox{0.94}{ 
    \begin{tabular}{|c|l|c|c|c|}
        \hline
        \textbf{$\sigma$} & \textbf{Method} & \textbf{$\hat{q}$ value} & \textbf{Avg.~length $\downarrow$}  & \textbf{Coverage(\%)}\\
        \hline
        \multirow{3}{*}{\large$\substack{\sigma_{\text{true}}=0.2 \\ 
        \hat{\sigma}=0.205}$} 
            & Oracle CP &  $2.33 \pm 0.02$& $0.66 \pm 0.01$ &$90.7 \pm 0.21$ \\
             & Noisy CP    & $3.27 \pm 0.04$  & $0.97 \pm 0.01$   & $95.7 \pm 0.34$    \\
            & Ours        & $2.33 \pm 0.02$ & $0.72 \pm 0.01$   &  $90.0 \pm 0.26$  \\
        \hline
        \multirow{3}{*}{\large$\substack{\sigma_{\text{true}}=0.3 \\ 
        \hat{\sigma}=0.314}$}  
           & Oracle CP &  $2.50 \pm 0.03 $& $0.77 \pm 0.01$ & $90.4 \pm 0.18$  \\
           & Noisy CP    & $3.92 \pm 0.05$  & $1.29 \pm 0.02$  &   $96.4 \pm 0.38$\\
            & Ours        & $2.52 \pm 0.02$ & $0.84 \pm 0.02$   &  $90.5 \pm 0.32$  \\
        \hline
    \end{tabular}
    }
    \label{tab:coverage_ci_noisy_training}
    \vspace{-3mm}
\end{table*}

\section{Experiments}
\label{sec:exp}

In this section, we evaluate the usefulness of our approach in reducing the confidence intervals while preserving coverage.

\noindent
\textbf{Datasets.} 
We use two medical imaging datasets. %
\textbf{BoneAge} \cite{halabi2019rsna}: hand CT age regression from the RSNA pediatric bone age dataset. \textbf{Chest X-Ray} \cite{nih_chestxray}:  large-scale collection of chest radiographs from the NIH, commonly used for disease classification tasks. The task in both datasets is to infer a person’s age from the images. We split the data into $60\%$ training, $30\%$ testing, and $10\%$ calibration.
We normalize the labels by the mean and standard deviation (SD) of the training labels.

\noindent
\textbf{Noise model.} 
We implement label noise in the calibration set (and in Section~\ref{seq:exp_noisy_training} also in the training set) prior to the label normalization by perturbing the ground truth ages with additive Gaussian noise and rounding them to integer units (months for \cite{halabi2019rsna}, years for \cite{nih_chestxray}).
We denote by $\sigma_{\text{true}}$ the SD of the label noise scaled by the empirical SD of the clean training labels. For example, $\sigma_{\text{true}}=0.2$ stands for using noise level that is fifth of the SD of the clean labels. %

\noindent
\textbf{Regression DNN.} 
Our model is based on EfficientNet-B4 \cite{tan2019efficientnet}, pretrained on ImageNet. We remove the final fully connected layer and add two task-specific heads: one for the mean prediction $\hat{y}$ and one for the log variance $\log \hat{u}^2$. 
We train the model with Gaussian NLL loss.

\noindent
\textbf{Compared methods.} 
We compare the following approaches: 
(1) \textit{Oracle CP}, which uses clean calibration labels\\
(2) \textit{Noisy CP}, which applies standard conformal prediction on noisy calibration set (motivated by \cite{einbinder2024label});\\ 
(3) \textit{Ours}, our CP method, %
as proposed in Section~\ref{sec:method}. \\
For all the methods we use $\alpha=0.1$ (as common in the literature), the score $s(x,y) = |y-\hat{y}(x)|/\hat{u}(x)$ and its decision rule $\mathcal{C}_q(\cdot)$.

\noindent
\textbf{Evaluation metrics.} 
We report the mean ($\pm$SD) of the CP threshold $\hat{q}$, the interval length and the coverage percentage (for clean test labels) computed over 6 trials.
Formally, the average interval length and the coverage rate are computed over the test set $\{(x_i^{(test)},y_i^{(test)})\}_{i=1}^{N_{test}}$ as follows:
    \begin{align*}
        &\text{Avg.~length} = \frac{1}{N_{test}} \sum_{i=1}^{N_{test}} |\mathcal{C}(x_i^{(test)})| \\
        &\hspace{17mm} =  \frac{1}{N_{test}} \sum_{i=1}^{N_{test}} 2 \hat{q} \hat{u}(x_i^{(test)}), \\
        &\text{Coverage rate} = \frac{1}{N_{test}} \sum_{i=1}^{N_{test}} \mathbf{1}\{ y_i \in \mathcal{C}(x_i^{(test)})\}.
    \end{align*}

\subsection{Noisy labels only in calibration}

We train a model on clean data and apply CP on calibration set with noisy labels generated with $\sigma_{\text{true}}=0.2$.
We apply our CP method with $\sigma=0.2$, but also with other values to assess its robustness to this parameter.

The results are reported in Table~\ref{tab:coverage_ci}.
As expected, for Noisy CP, $\hat{q}$ is significantly larger than for Oracle CP. Accordingly, it suffers from large confidence intervals and over-coverage.
In contrast, our method with correct $\sigma$ reaches near-oracle performance, with intervals much smaller than Noisy CP while providing the target coverage.
We see that applying our method with a smaller $\sigma$ (can be interpreted as: not addressing some of the noise) yields over-coverage and larger intervals (similar to the relation between Noisy CP and Oracle CP), and vice versa for a larger $\sigma$. However, the changes in the metrics for our method are rather minor, which demonstrates its robustness.

\subsection{Noisy labels in both training and calibration}
\label{seq:exp_noisy_training}

We consider label noise of level $\sigma_{\text{true}}$ in both the training set and the calibration set.
We use the procedure described in Section~\ref{seq:noise_est} to get $\hat{\sigma}$, an estimate of $\sigma_{\text{true}}$, for applying our CP method.
We examine $\sigma_{\text{true}} \in \{0.2, 0.3\}$.

The results are reported in Table~\ref{tab:coverage_ci_noisy_training}.
For serving as a meaningful performance bound, the Oracle CP also uses the same models that are trained with noisy labels (but has a clean calibration set). This explains the slight degradation in its performance compared to Table~\ref{tab:coverage_ci_noisy_training} for $\sigma_{\text{true}}=0.2$.
As expected, the increase in the intervals of Noisy CP becomes more significant as the noise level increases.
Aligned with the robustness property demonstrated above, our method benefits from the fact that $\hat{\sigma}$ is close to $\sigma_{\text{true}}$ and demonstrates near oracle performance, significantly outperforming Noisy CP.

\section{Conclusion}
\label{sec:conclusion}

We addressed the problem of applying CP for regression DNNs with noisy labels in the calibration set.
We developed an iterative deconvolution-based procedure for estimating the noise-free CP threshold, based on a solid mathematical derivation.
We evaluated the method on two medical imaging regression datasets with Gaussian label noise and showed that it outperforms the existing alternative and achieves performance close to the clean-label setting.

%
\bibliographystyle{IEEEbib}
\bibliography{refs,copa_ref,midl25}

\end{document}